\documentclass{article}

\PassOptionsToPackage{numbers, sort&compress}{natbib}

\usepackage[preprint]{neurips_2023}

\usepackage{amsmath,amsfonts,bm}

\def\1{\bm{1}}

\def\vzero{{\bm{0}}}

\def\vmu{{\bm{\mu}}}

\def\vepsilon{{\bm{\epsilon}}}

\def\vx{{\bm{x}}}

\def\vz{{\bm{z}}}

\def\mI{{\bm{I}}}

\DeclareMathAlphabet{\mathsfit}{\encodingdefault}{\sfdefault}{m}{sl}
\SetMathAlphabet{\mathsfit}{bold}{\encodingdefault}{\sfdefault}{bx}{n}

\def\gD{{\mathcal{D}}}
\def\gE{{\mathcal{E}}}

\def\gN{{\mathcal{N}}}

\def\sR{{\mathbb{R}}}

\newcommand{\E}{\mathbb{E}}

\usepackage[utf8]{inputenc} 
\usepackage[T1]{fontenc}    
\usepackage{hyperref}       
\usepackage{url}            
\usepackage{booktabs}       
\usepackage{amsfonts}       
\usepackage{nicefrac}       
\usepackage{microtype}      
\usepackage{xcolor}         
\usepackage{amsmath,amssymb}
\usepackage{graphicx}
\usepackage{subfig}
\usepackage{mathtools}
\usepackage{algorithm}
\usepackage{algorithmic}
\usepackage{amsthm}
\usepackage{booktabs}  
\usepackage{arydshln}
\usepackage{subfig}
\usepackage{multirow}

\title{ControlVideo: Conditional Control for One-shot Text-driven Video Editing and Beyond}

\author{%
  Min Zhao$^{1,3}$, Rongzhen Wang$^{2}$, Fan Bao$^{1,3}$, Chongxuan Li$^{2}$\thanks{The Corresponding authors.}~~, Jun Zhu$^{1,3,4 *}$\\
  $^1$Dept. of Comp. Sci. $\&$ Tech., BNRist Center, THU-Bosch ML Center, Tsinghua University, China\\
  $^2$ Gaoling School of Artificial Intelligence, Renmin University of China, Beijing, China\\
  Beijing Key Laboratory of Big Data Management and Analysis Methods , Beijing, China\\
$^3$ShengShu, Beijing, China;~~$^4$Pazhou Laboratory (Huangpu), Guangzhou, China \\
  \texttt{gracezhao1997@gmail.com;} \texttt{wangrz@ruc.edu.cn;}
  \texttt{bf19@mails.tsinghua.edu.cn;} \\
  \texttt{chongxuanli@ruc.edu.cn;} \texttt{dcszj@tsinghua.edu.cn} 
}

\begin{document}

\maketitle
\begin{abstract}
This paper presents \emph{ControlVideo} for text-driven video editing -- generating a video that aligns with a given text while preserving the structure of the source video. Building on a pre-trained text-to-image diffusion model, ControlVideo enhances the fidelity and temporal consistency by incorporating additional conditions (such as edge maps), and fine-tuning the key-frame and temporal attention on the source video-text pair via an in-depth exploration of the design space. Extensive experimental results demonstrate that ControlVideo outperforms various competitive baselines by delivering videos that exhibit high fidelity w.r.t. the source content, and temporal consistency, all while aligning with the text. By incorporating Low-rank adaptation layers into the model before training, ControlVideo is further empowered to generate videos that align seamlessly with reference images. More importantly, ControlVideo can be readily extended to the more challenging task of long video editing (e.g., with hundreds of frames), where maintaining long-range temporal consistency is crucial. To achieve this, we propose to construct a fused ControlVideo by applying basic ControlVideo to overlapping short video segments and key frame videos and then merging them by pre-defined weight functions. Empirical results validate its capability to create videos across 140 frames, which is approximately 5.83 to 17.5 times more than what previous works achieved. The code is available at \href{https://github.com/thu-ml/controlvideo}{https://github.com/thu-ml/controlvideo} and the visualization results are available at \href{https://drive.google.com/file/d/1wEgc2io3UwmoC5vTPbkccFvTkwVqsZlK/view?usp=drive_link}{HERE}.

\end{abstract}

\section{Introduction}

The endeavor of text-driven video editing is to generate videos derived from textual prompts and existing video footage, thereby reducing manual labor. This technology stands to significantly influence an array of fields such as advertising, marketing, and social media content. During this process, it is critical for the edited videos to \emph{faithfully} preserve the content of the source video, maintain \emph{temporal consistency} between generated frames, and \emph{align} with the provided text and optional reference images. However, fulfilling all these requirements simultaneously poses substantial challenges. What's more, a further challenge arises when dealing with real-world videos that typically consist of hundreds of frames: how can \emph{long-range temporal consistency} be maintained? 

\label{sec: introduction}

Previous research~\cite{qi2023fatezero,wang2023zero,wu2022tune,liu2023video} has made significant strides in text-driven video editing under zero-shot and one-shot settings, capitalizing on advancements in large-scale text-to-image (T2I) diffusion models ~\cite{rombach2022high,ho2022imagen} and image editing techniques~\cite{hertz2022prompt,tumanyan2022plug,parmar2023zero}. However, despite these advancements, they still cannot address the aforementioned challenges. \emph{First}, empirical evidence (see Fig.~\ref{fig:baseline}) suggests that existing approaches still struggle with fulfilling three requirements of text-driven video editing simultaneously, such as faithfully controlling the output while preserving temporal consistency.  \emph{Second}, these approaches primarily focus on short video editing, specifically videos shorter than 24 frames, and do not explore how to maintain temporal consistency over extended durations.

To address the first challenge, we present \emph{ControlVideo} for faithful and temporal consistent video editing, building upon a pre-trained T2I diffusion model. To enhance fidelity, we propose to incorporate visual conditions such as edge maps as additional inputs into T2I diffusion models to amplify the guidance from the source video. As ControlNet \cite{zhang2023adding} has been pre-trained alongside the diffusion model, we utilize it to process these visual conditions. Recognizing that various visual conditions encompass varying degrees of information from the source video, we engage in a comprehensive investigation of the suitability of different visual conditions for different scenes. This exploration naturally leads us to combine multiple controls to leverage their respective advantages. Furthermore, we transform the original spatial self-attention into key-frame attention, aligning all frames with a selected one, and incorporate temporal attention modules as extra branches in the diffusion model to improve faithfulness and temporal consistency further, which is designed by a systematic empirical study. Additionally, ControlVideo can generate videos that align with optional reference images by introducing Low-rank adaptation (LoRA)~\cite{hu2021lora} layers on the diffusion model before training. 

\begin{figure}
\centering
\includegraphics[width=1.0\columnwidth]{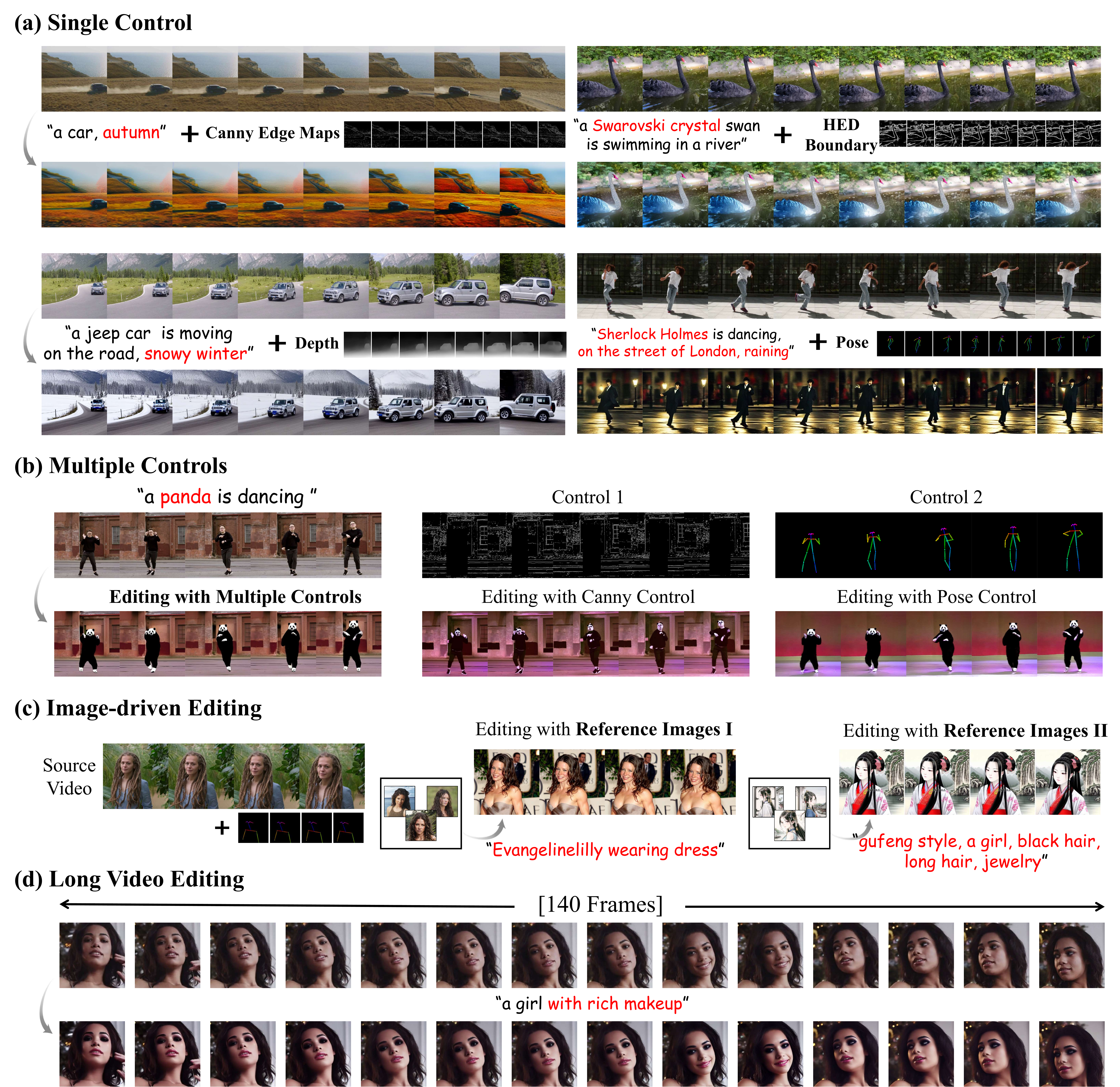}
  \caption{Main results of ControlVideo with (a) single control, (b) multiple controls, (c) image-driven video editing, and (d) long video editing.}
   \label{fig: main results}
   \vspace{-0.2cm}
\end{figure}

Empirically,  we validate our method on 50 video-text pair data collected from the Davis dataset following previous works~\cite{qi2023fatezero,liu2023video,wu2022tune} and the internet. We compare with Stable Diffusion and SOTA text-driven video editing methods~\cite{qi2023fatezero,liu2023video,wu2022tune} under objective metrics and a user study.  In particular, following~\cite{qi2023fatezero,liu2023video} we use CLIP~\cite{radford2021learning} to measure text-alignment and temporal consistency and employ SSIM to assess faithfulness. Extensive results demonstrate that ControlVideo outperforms various competitors by fulfilling three requirements of text-driven video editing simultaneously. Notably, ControlVideo can produce videos with extremely realistic visual quality and very faithfully preserve original source content while following the text guidance. For instance, ControlVideo can successfully make up a woman with maintaining her identity while all existing methods fail (see Fig. \ref{fig:baseline}).  

Furthermore, ControlVideo is readily extendable for the aforementioned second challenge: video editing for long videos that encompass hundreds of frames (see Sec. \ref{sec: extended controlvideo}). To achieve this, we propose to construct a fused ControlVideo by applying basic ControlVideo to overlapping short videos and key frame videos and then merging them by defined weight functions at each denoising step. Intuitively, fusion with overlapping short videos encourages the overlapping frames to merge features from neighboring short videos, thereby effectively mitigating inconsistency issues between adjacent video clips. On the other hand, key frame video, which incorporates the first frame of each video segment, provides global guidance from the whole video, and thus fusion with it can further improve long-range temporal consistency. Empirical results affirm ControlVideo's ability to produce videos spanning 140 frames, which is approximately 5.83 to 17.5 times longer than what previous works handled.

\section{Background}

\subsection{Diffusion Models for Image Generation and Editing} 

Let $q(\vx_0)$ be the data distribution on $\sR^D$. Diffusion models~\cite{song2020score,bao2021analytic,ho2020denoising} gradually perturb data $\vx_0 \sim q(\vx_0)$ by a forward diffusion process:
\begin{align}
    q(\vx_{1:T}) = q(\vx_0)\prod_{t=1}^T q(\vx_t | \vx_{t-1}), \quad q(\vx_t | \vx_{t-1}) = \gN(\vx_t;\sqrt{\alpha_t}\vx_{t-1},\beta_t \mI),
\end{align}
where $\beta_t$ is the noise schedule, $\alpha_t = 1 - \beta_t$ and is designed to satisfy $\vx_T \sim \gN(\vzero, \mI)$. The forward process $\{\vx_t\}_{t \in [0,T]}$ has the following transition distribution: 
\begin{align}
\label{eq: forward transition distribution}
    q_{t|0}(\vx_t|\vx_0) = \gN(\vx_t | \sqrt{\Bar{\alpha}_t} \vx_0,(1-\Bar{\alpha}_t)\mI ),
\end{align}
where $\Bar{\alpha}_t=\prod_{s=1}^t \alpha_s$. The data can be generated starting from $\vx_T \sim \gN(\vzero, \mI)$ through the reverse diffusion process, where the reverse transition kernel $q(\vx_{t-1}|\vx_t)$ is learned by a Gaussian model: $ p_{\theta}(\vx_{t-1}|\vx_t) = \gN(\vx_{t-1};\vmu_{\theta}(\vx_t),\sigma_t^2 \mI)$. Ho et al.~\cite{ho2020denoising} shows learning the mean $\vmu_{\theta}(\vx_t)$ can be derived to learn a noise prediction network $\vepsilon_\theta(\vx_t, t)$ via a mean-squared error loss:
\begin{align}
    \label{eq: training objective}
    \min_{\theta} \E_{t, \vx_0, \vepsilon}||\vepsilon - \epsilon_\theta(\vx_t, t)||^2,
\end{align}
where $\vx_t\sim q_{t|0}(\vx_t|\vx_0), \vepsilon\sim \gN(\vzero,\mI)$. Deterministic DDIM sampling~\cite{song2020denoising} generate samples starting from $\vx_T \sim \gN(\vzero, \mI)$ via the following iteration rule:
\begin{align}
    \label{eq: sampling}
    \vx_{t-1} = \sqrt{\alpha_{t-1}}\frac{\vx_t - \sqrt{1-\alpha_t}\vepsilon_\theta(\vx_t, t)}{\sqrt{\alpha_t}} + \sqrt{1-\alpha_{t-1}}\vepsilon_\theta(\vx_t, t).
\end{align}
Due to the ability to generate high-quality samples, diffusion models are naturally applied to in image translation and image editing~\cite{meng2021sdedit,zhao2022egsde,hertz2022prompt}. Unlike unconditional generation, they usually need to preserve the content from the source image $\vx_0$. Considering the reversible property of ODE, DDIM inversion~\cite{song2020denoising} is adopted to convert a real image $\vx_0$ to related inversion noise $\vx_M$ by reversing the above process for faithful image editing:
\begin{align}
    \label{eq: ode inversion}
    \vx_{t} = \sqrt{\alpha_{t}}\frac{\vx_{t-1} - \sqrt{1-\alpha_{t-1}}\vepsilon_\theta(\vx_{t-1}, t-1)}{\sqrt{\alpha_{t-1}}} + \sqrt{1-\alpha_{t}}\vepsilon_\theta(\vx_{t-1}, t-1).
\end{align}

\subsection{Latent Diffusion Models and ControlNet} To reduce computational cost, latent diffusion models (LDM, a.k.a Stable Diffusion)~\cite{rombach2022high} use an encoder $\gE$ to transform $\vx_0$ into low-dimensional latent space $\vz_0 = \gE(\vx_0)$, which can be reconstructed by a decoder $\vx_0 \approx \gD(\vz_0)$, and then learns the noise prediction network $\epsilon_\theta(\vz_t, p, t)$ in the latent space, where $p$ is the textual prompts. The backbone for $\epsilon_\theta(\vz_t, p, t)$ is the UNet (termed \emph{main UNet}) that stacks several basic blocks. Specifically, the U-Net consists of an encoder, a middle block, and a decoder. The encoder and decoder each consist of 12 blocks, while the full model encompasses a total of 25 blocks. Within these blocks, 8 are utilized for down-sampling or up-sampling convolution layers, and the remaining blocks constitute the basic building blocks. Each basic block is composed of a transformer block and a residual block. The transformer block incorporates a self-attention layer, a cross-attention layer, and a feedforward neural network. The text embeddings, processed by CLIP text encoder, are integrated into the U-Net via the cross-attention layer. To enable models to learn additional conditions $c$, {ControlNet~\cite{zhang2023adding} adds a trainable copy of the encoder and middle blocks of the main UNet (termed \emph{ControlNet}) to incorporate task-specific conditions on the locked Stable Diffusion. The outputs of ControlNet are then followed by a zero-initialization convolutional layer, which is subsequently added to the features of the main U-Net at the corresponding layer.

\section{Methods}
\label{pa:controlvideo}
 To address the challenges mentioned in Sec. \ref{sec: introduction}, we first present ControlVideo for faithful and temporally consistent text-driven video editing building upon a pre-trained T2I diffusion model (see Sec. \ref{sec: ControlVideo}). Then we extend ControlVideo for the second challenge: video editing for long videos that encompass hundreds of frames (see Sec. \ref{sec: extended controlvideo}).  

\subsection{ControlVideo}
\label{sec: ControlVideo}
In this section, we first introduce the architecture of ControlVideo via an in-depth exploration of the design space (see Sec. \ref{sec: architecture}). As shown in Figure \ref{fig:method}, ControlVideo incorporates additional conditions, fine-tuning the key-frame, and temporal attention. In Sec. \ref{sec: training and sampling framework}, we present the training and sampling framework of ControlVideo. Furthermore, we show how ControlVideo can produce videos in alignment with optional reference images by incorporating Low-rank adaptation layers in Sec. \ref{sec: image-driven}.

\subsubsection{Architecture}
\label{sec: architecture}
As the T2I diffusion model has been pre-trained on large-scale text-image data, we build upon it to align with given texts. In line with prior studies~\cite{wu2022tune,qi2023fatezero}, we first replace the spatial kernel ($3\times3$ ) in 2D convolution layers with 3D kernel ($1\times3\times3$) to handle videos inputs.

\textbf{Adding Visual Controls.} Recall that a key objective in text-driven video editing is to \emph{faithfully} preserve the content of the source video. An intuitive approach is to generate edited videos starting from DDIM inversion $X_M$ in Eq. \ref{eq: ode inversion} to leverage information from $X_0$. However, despite the reversible nature of ODE, as depicted in Fig.~\ref{fig:ablations}, empirically, the combination of DDIM inversion and DDIM sampling significantly disrupts the structure of the source video. To enhance fidelity, we propose to introduce additional visual conditions $C=\{c^i\}_{i=1}^N$, such as edge maps for all frames, into the main UNet to amplify the source video's guidance at each time step rather than only initial time: $\epsilon_\theta(X_t, C, p, t)$. Notably, as ControlNet\cite{zhang2023adding} has been pre-trained alongside the main UNet in Stable Diffusion, we utilize it to process these visual conditions $C$. Formally, let $h_u \in \sR^{N \times d}$ and $h_c \in \sR^{N \times d}$ denote the hidden features with dimension $d$ of the same layer in the main UNet and ControlNet, respectively. We combine these features by summation, yielding $h = h_u + \lambda h_c$, which is then fed into the decoder of the main UNet through a skip connection, with $\lambda$ serving as the control scale. As illustrated in Figure~\ref{fig:ablations}, the introduction of visual conditions to provide structural guidance from $X_0$ significantly enhances the faithfulness of the edited videos. 

Further, given that different visual conditions encompass varying degrees of information derived from $X_0$, we comprehensively investigate \emph{the advantages of employing different conditions}. As depicted in Figure \ref{fig: main results}, our findings indicate that conditions yielding detailed insights into $X_0$, such as edge maps, are particularly advantageous for attribute manipulation such as facial video editing, demanding precise control to preserve human identity. Conversely, conditions offering coarser insights into $X_0$, such as pose information, facilitate flexible adjustments to shape and background. This exploration naturally raises the question of whether we can combine \emph{multiple controls} to leverage their respective advantages. To this end, we compute a weighted sum of hidden features derived from different controls, denoted as $h = h_u + \sum_{i} \lambda_i h_c$, and subsequently feed the fused features into the decoder of the main UNet, where $\lambda_i$ represents the control scale associated with the $i$-th control. In situations where multiple controls may exhibit conflicts or inconsistencies, we can employ  SAM~\cite{kirillov2023segany} or cross-attention map~\cite{hertz2022prompt} to generate a mask based on text and feed the masked controls into ControlVideo to enhance control synergy. As shown in Figure \ref{fig: main results}, Canny edge maps excel at preserving the background while having a limited impact on shape modification. In contrast, pose control facilitates flexible shape adjustments but may overlook other crucial details. By combining these controls, we can simultaneously preserve the background and effect shape modifications, demonstrating the feasibility of leveraging multiple controls in complex video editing scenarios.

\begin{figure}
  \centering
  \includegraphics[width=1\columnwidth]{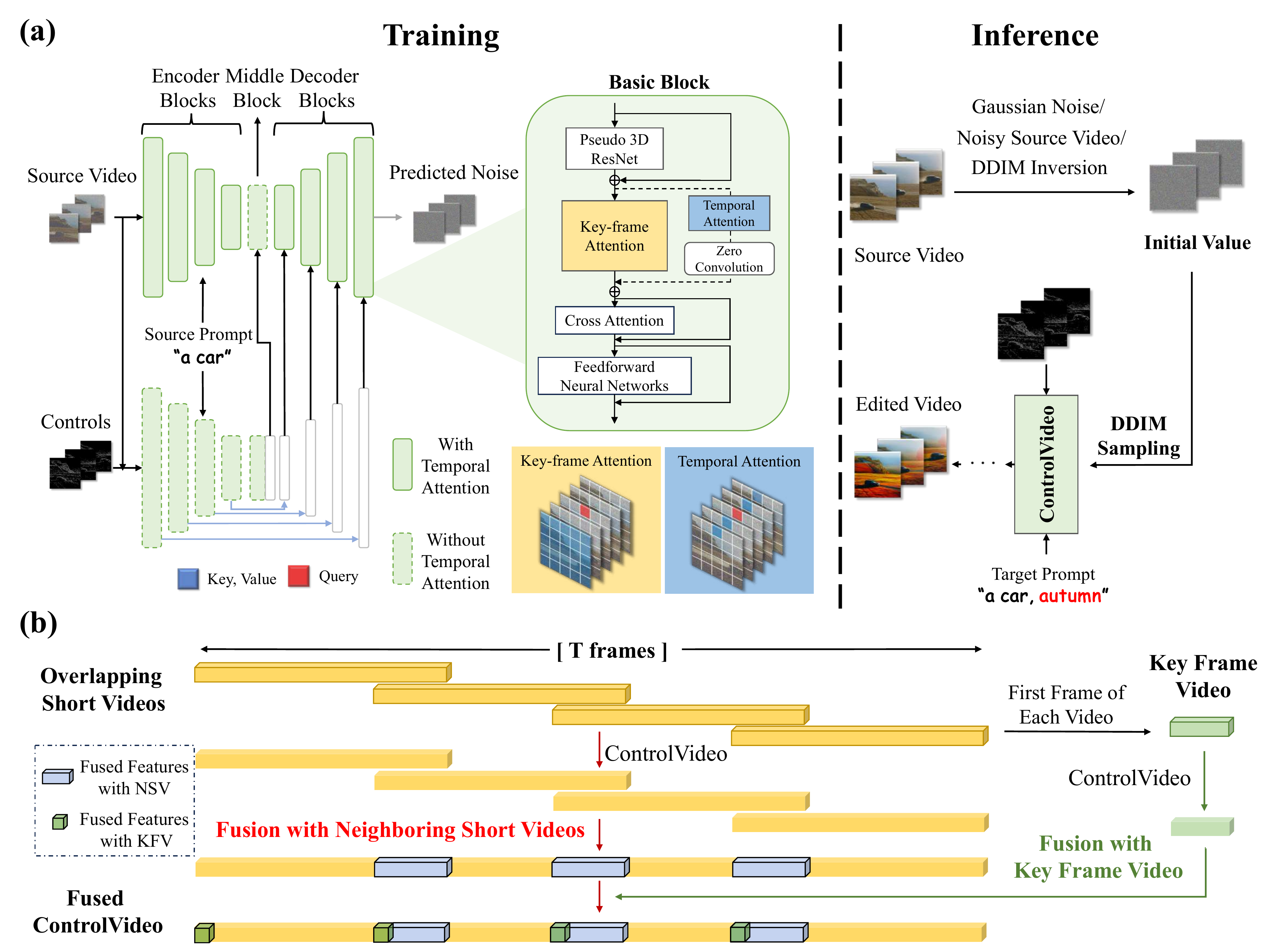}
\caption{(a) The overview of ControlVideo. \textbf{Left}: the architecture. ControlVideo incorporates additional controls, fine-tunes the key-frame attention, and temporal attention. The attention modules are initialized using the self-attention weights from T2I diffusion models. \textbf{Right}: the inference framework. Depending on the editing scenarios, we have three ways to derive initial values (see Sec. \ref{sec: training and sampling framework}). (b) The overview of extended ControlVideo for long video editing. NSV and KFV represent neighboring short videos and key frame videos respectively.}
  \label{fig:method}
  \vspace{-0.2cm}
\end{figure}

\textbf{Key-frame Attention.} 
The T2I diffusion models update the features of each frame independently and have no interaction between frames, thus resulting in temporal inconsistencies. To address this issue and improve \emph{temporal consistency}, we introduce a key frame that serves as a reference for propagating information throughout the video. Specifically, drawing inspiration from previous works~\cite{wu2022tune}, we transform the spatial self-attention in both main UNet and ControlNet into key-frame attention, aligning all frames with a selected reference frame. Formally, let $v^i\in\sR^{d}$ represent the hidden features of the $i$-th frame, and let $k\in[1, N]$ denote the chosen key frame. The key-frame attention mechanism is defined as follows:
\begin{align*}
    Q = W^Q v^i , K = W^K v^k , V = W^V v^k,
\end{align*}
where $W^Q, W^K, W^V$ are the projected matrix. We initialize these matrices using the original self-attention weights to leverage the capabilities of T2I diffusion models fully. Empirically, we systematically study \emph{the design of key frame, key and value selection in self-attention and fine-tuned parameters}. A detailed analysis is provided in Appendix. In summary, we utilize the first frame as key frame, which serves as both the key and value in the attention mechanism, and we finetune the output projected matrix $W^O$ within the attention modules to enhance temporal consistency.

\begin{figure}
  \centering
  \includegraphics[width=1.0\columnwidth]{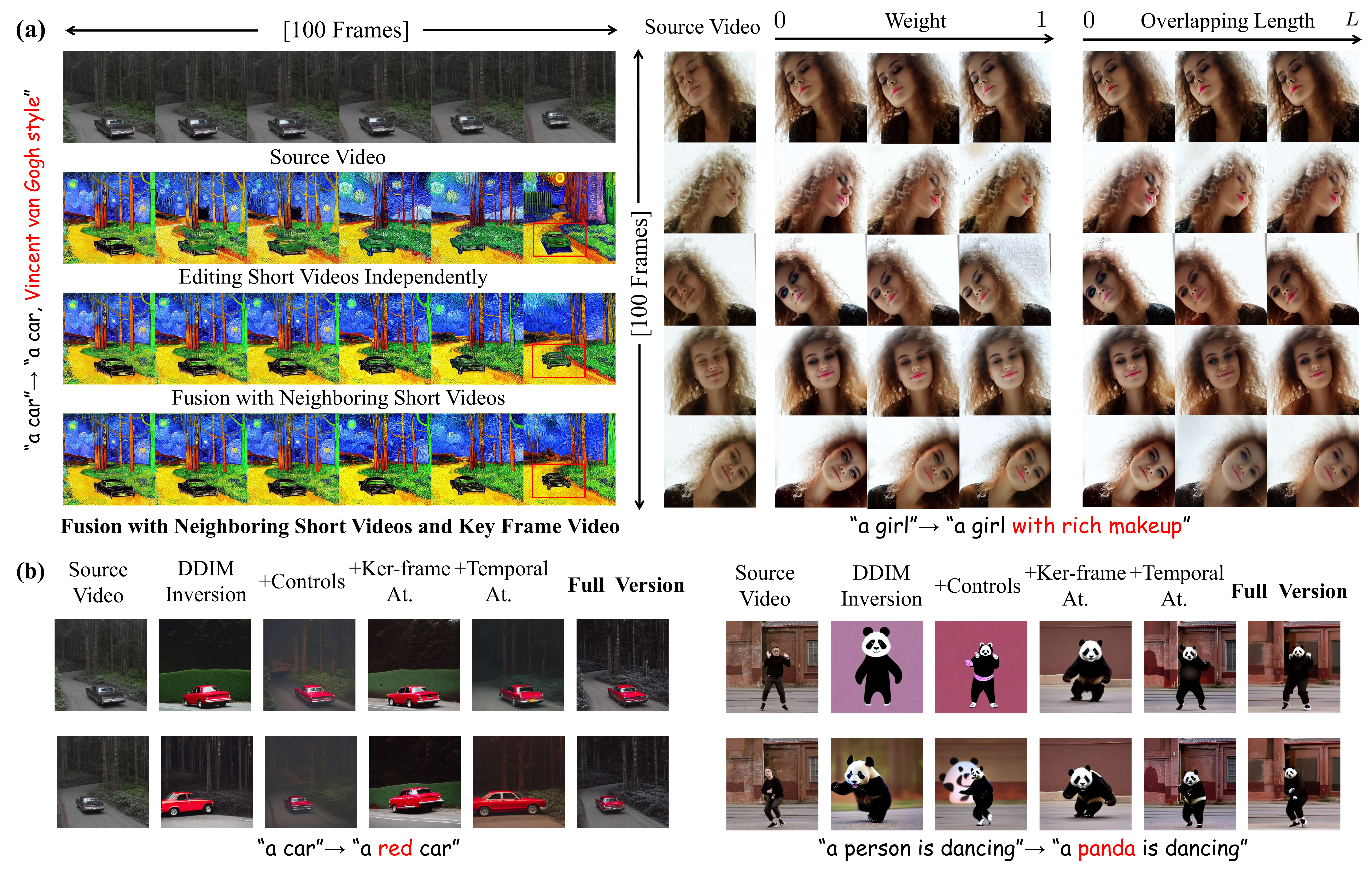}
  \caption{(a) Ablation study for fusion strategies, overlapping length $a$ and weight $w$ for key frame video fusion for long video editing.  See detailed analysis in Sec. \ref{sec: extended controlvideo} and Sec. \ref{sec: ablations}. (b) Ablation studies for key components in ControlVideo. At. denote attention. See detailed analysis in Sec. \ref{sec: ablations}.}
  \label{fig:ablations}
\end{figure}

\textbf{Temporal Attention.} 
In pursuit of enhancing both the \emph{faithfulness} and \emph{temporal consistency} of the edited video, we introduce temporal attention modules as extra branches in the network, which capture relationships among corresponding spatial locations across all frames. Formally, let $v \in \mathbb{R}^{N\times d}$ denote the hidden features, the temporal attention is defined as follows:
\begin{align*}
    Q = W^Q v , K = W^K v , V = W^V v.
\end{align*} 
Prior research~\cite{singer2022make} has benefited from extensive data to train temporal attention, a luxury we do not have in our one-shot setting. To address this challenge, we draw inspiration from the consistent manner in which different attention mechanisms model relationships between image features. Accordingly, we initialize temporal attention using the original spatial self-attention weights, harnessing the capabilities of the T2I diffusion model. After each temporal attention module, we incorporate a zero convolutional layer~\cite{zhang2023adding} to retain the module's output prior before fine-tuning. Furthermore, we conduct a comprehensive study on \emph{the incorporation of local and global positions for introducing temporal attention}. The qualitative results are shown in Figure \ref{fig:temporal}. Concerning local positions in the transformer block, we find that the most effective placement is both before and within the self-attention mechanism. This choice is substantiated by the fact that the input in these two positions matches that of self-attention, serving as the initial weight for temporal attention. With self-attention location exhibits higher text alignment, ultimately making it our preferred choice. For global location in ControlVideo, our main finding is that the effectiveness of positions is correlated with the amount of information they encapsulate. For instance, the main UNet responsible for image generation retains a full spectrum of information, outperforming the ControlNet, which focuses solely on extracting condition-related features while discarding others. As a result, we incorporate temporal attention alongside self-attention at all stages of the main UNet, with the exception of the middle block. More detailed analyses are provided in Appendix. 

\subsubsection{Training and Sampling Framework}
\label{sec: training and sampling framework}

 Let $C=\{c^i\}_{i=1}^N$ denote the visual conditions (e.g., Canny edge maps) for $X_0$ and $\epsilon_{\theta}(X_t, C,p,t)$ denote the ControlVideo network. Let $p_s$ and $p_t$ represent the source prompt and target prompt, respectively. Similar to Eq. \ref{eq: training objective}, we finetune $\epsilon_{\theta}(X_t,C,p,t)$  on the source video-text pair $(X_0, p_s)$ using the mean-squared error loss, defined as follows:
\begin{align*}
    \min_{\theta} \E_{t, \vepsilon }||\vepsilon - \vepsilon_{\theta}(X_t,C, p_s, t)||^2,
\end{align*}
where $\vepsilon \sim \gN(\vzero, \mI), X_t \sim q_{t|0}(X_t|X_0)$. Note that during training, we exclusively optimize the parameters within the attention modules (as discussed in Sec. \ref{sec: architecture}), while keeping all other parameters fixed.

\textbf{Choice of Initial Value $X_M$.} Built upon $\vepsilon_{\theta}(X_t, C, p, t)$, we can generate the edited video starting from the initial value $X_M$ using DDIM sampling~\cite{song2020denoising}, based on the target prompt $p_t$. For $X_M$, we employ DDIM inversion as described in Eq. \ref{eq: ode inversion} for local editing tasks, such as attribute manipulation. For global editing, different from previous work~\cite{wu2022tune,qi2023fatezero}, we can also start from noisy source video $X_M\sim q_{M|0}(X_M|X_0)$ using forward transition distribution in Eq. \ref{eq: forward transition distribution} with large $M$ and even $X_M\sim \gN(\vzero,\mI)$  to improve editability because visual conditions have already provided structure guidance from $X_0$. During this process, the sampled noise is shared across all frames for temporal consistency.

\begin{algorithm}
    \caption{Extended ControlVideo for Long Video Editing}
    \label{alg: long video editing}
    \begin{algorithmic}
        \REQUIRE initial value $X_M$, controls $C$, short video length $L$, overlapped length $a$, fusion function $F(\cdot)$, weight $w$, model $\vepsilon_\theta(\cdot,\cdot,\cdot,\cdot)$, prompt $p$  
        \STATE $n=\lfloor N/(L-a)\rfloor + 1$  \hfill $\blacktriangleright$ number of short videos
    \FOR{$t = M$ to $1$}
        \FOR{$j = 1$ to $n$}
        \STATE $ \vepsilon_\theta^j \leftarrow \vepsilon_\theta(X_{t}^j,C^j, p, t)$ \hfill $\blacktriangleright$ ControlVideo for each short video
        \ENDFOR
        \STATE $\hat{\vepsilon}_\theta \leftarrow F(\vepsilon_\theta^1, \dots, \vepsilon_\theta^n)$ \hfill $\blacktriangleright$ fusion with neighboring short videos via Eq. \ref{eq: fusion function}
        \STATE $\epsilon_\theta^K \leftarrow \vepsilon_\theta(X_t^K, C^K, p, t)$ \hfill $\blacktriangleright$ ControlVideo for key frame video
        \STATE $\vepsilon_\theta \leftarrow w O(\epsilon_\theta^K) + (1-w)\hat{\vepsilon}_\theta$\hfill $\blacktriangleright$ fusion with key frame video via Eq. \ref{eq:  key fusion}
        \STATE $X_{t-1} \leftarrow$ DDIM$\_$Sampling($\vepsilon_\theta, X_{t}, t$) \hfill $\blacktriangleright$ denoising step in Eq. \ref{eq: sampling}
        
    \ENDFOR
    \RETURN $X_{0}$
    \end{algorithmic}
\end{algorithm}

\subsubsection{Image-driven Video Editing} 
\label{sec: image-driven}

In certain scenarios, textual descriptions may fall short of fully conveying the precise desired effects from users. In such cases, users may wish for the generated video to also \emph{align} with given reference images. Here, we show a simple way to extend ControlVideo for image-driven video editing. Specifically, we can first add the Low-rank adaptation (LoRA)\cite{hu2021lora} layer on the main UNet to facilitate the learning of concepts relevant to reference images and then freeze them to train ControlVideo following Sec. \ref{sec: training and sampling framework}. Since the training for reference images and video is independent, we can flexibly utilize models in the community like CivitAI.

\begin{figure}
  \centering
  \includegraphics[width=1.0\columnwidth]{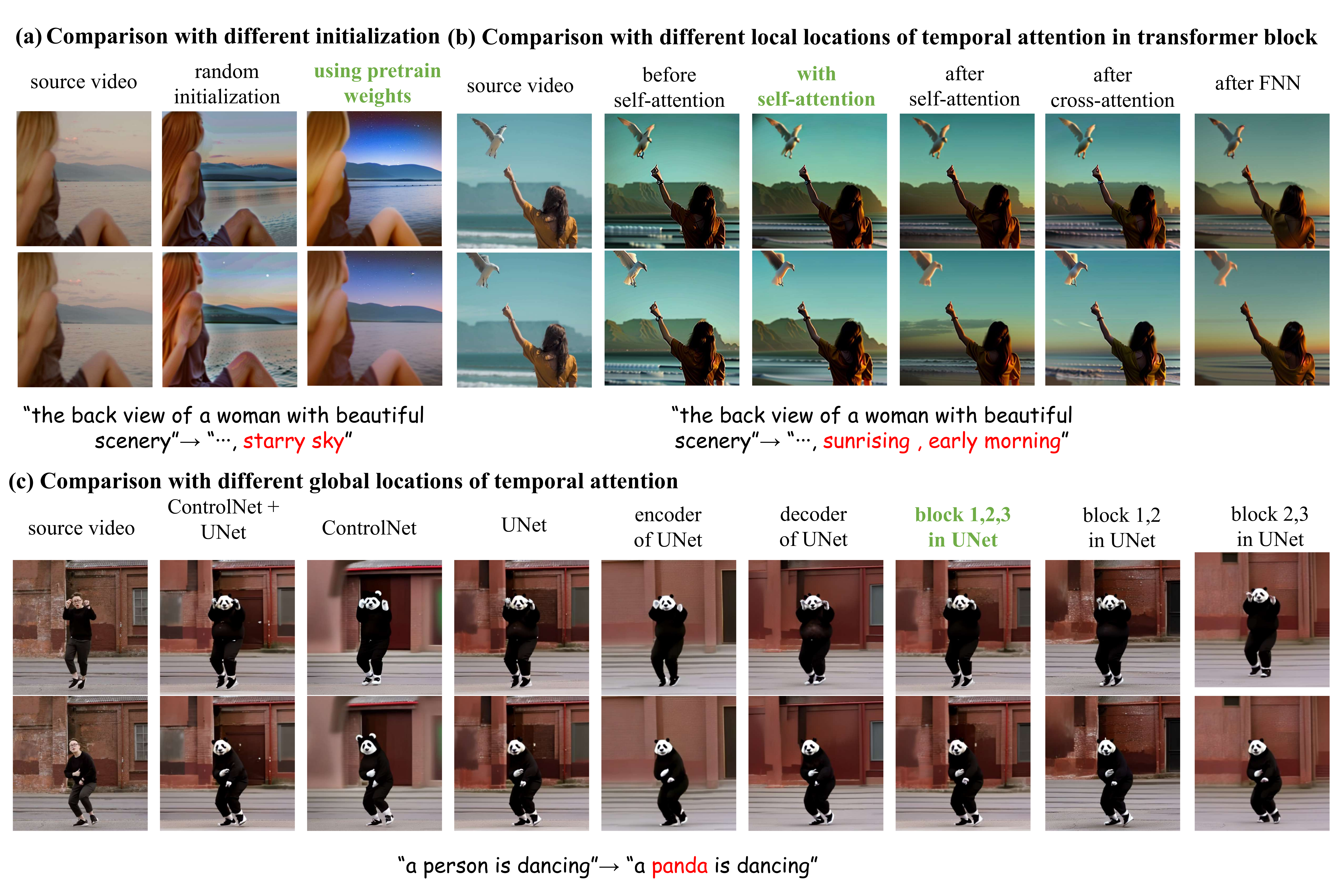}
  \caption{Ablation studies of (a) the way to initialize and the incorporation of (b) local positions and (c) global positions for introducing temporal attention. The green color marked our choice.}
  \label{fig:temporal}
\end{figure}

\subsection{Extended ControlVideo for Long Video Editing}
\label{sec: extended controlvideo}
 Although ControlVideo described in the above section has the appealing ability to generate highly temporal consistent videos, it is still difficult to deal with real-world videos that typically encompass hundreds of frames due to memory limitations. A straightforward approach to address this issue involves dividing the entire video into several non-overlapping short segments and applying ControlVideo to each segment independently. However, as depicted in Figure \ref{fig:ablations}, this method still results in temporal inconsistencies between video clips. To tackle this problem, we propose to fuse the features of the frames that bridge between short videos at each denoising step. To achieve this, as shown in Figure \ref{fig:method}, we split the whole video into overlapping short videos, apply ControlVideo for each segment, and
then merge features of overlapping frames from neighboring short videos via pre-defined weight functions, where the weight fusion strategy is also used in the image generation task~\cite{jimenez2023mixture}. Furthermore, in the subsequent denoising step, both non-overlapping and overlapping frames within a short video clip are fed into ControlVideo together, which brings the features of non-overlapping frames closer to those of the overlapping frames, thus indirectly improving global temporal consistency. Formally, the $j$-th short video clip $X_t^j$ and the corresponding visual conditions $C^j$ are defined as:
\begin{align}
\label{eq: split long video}
    X_t^j = \{\vx_t^i\}_{i={(j-1)(L-a)+1}}^{\min((j-1)(L-a) + L, N)}, \quad C^j = \{c^i\}_{i={(j-1)(L-a)+1}}^{\min((j-1)(L-a)+L, N)}, \quad j \in [1, n]
\end{align}
where $n=\lfloor N/(L-a)\rfloor + 1$ is the number of short video clips, $L$ is the length of short video clip and $a$ is the overlapped length.  Let $ \vepsilon_\theta^j\in \sR^{L\times D} = \vepsilon_\theta(X_{t}^j, C^j, p, t) $ denote the ControlVideo for $j$-th short video and $\hat{\vepsilon}_\theta\in \sR^{N \times D}$ denote the fused ControlVideo for entire video. The fusion function $F(\cdot):\sR^{n\times L\times D} \to \sR^{N\times D}$ is defined as follows: 
\begin{align}
    \label{eq: fusion function}
    \hat{\vepsilon}_\theta = F(\vepsilon_\theta^1, \dots, \vepsilon_\theta^n)= \textrm{Sum}(\textrm{Normalize} (O(w_j\otimes \mathbf{1}_D)) \odot O(\vepsilon_\theta^j)),
\end{align}
where $w_j\in \sR_{+}^{L}$ is the weight vector for the $j$-th short video, $\mathbf{1}_D\in \sR^D$ is a vector of ones, $\otimes$ is vector outer product, $\odot$ is the element-wise multiplication and Sum$(\cdot)$ adds elements at corresponding positions in the matrix. $O(\cdot): \sR^{L\times D} \to \sR^{N\times D}$ denote zero-padding. For instance, $O(\vepsilon_\theta^j)$ represents the corresponding frame indexes of $j$-th video are $\vepsilon_\theta^j$ and the other frame indexes are zero. Normalize$(\cdot)$ scales matrix elements by their sum at corresponding positions, ensuring fusion weights sum to one and maintaining value range post-fusion. In this work, we define normal random variables $w_j \sim \mathcal{N}(l; L/2,\sigma^2)$, where $\sigma=0.1$. Alternative weight functions were tested, with results indicating insensitivity to the choice of function (see Sec. \ref{sec: ablations for long video}). As shown in Figure \ref{fig:ablations}, this fusion strategy significantly enhances temporal consistency between short videos. 

However, this approach directly fuses nearby videos to ensure local consistency between adjacent video clips, and global consistency for the entire video is improved indirectly during repeated denoising steps. Consequently, as illustrated in Figure \ref{fig:ablations}, temporal consistency deteriorates when video clips are spaced farther apart, exemplified by the degradation of the black car into the green car. In light of these observations, a natural question arises: can we fuse more global features directly to enhance long-range temporal consistency further? To achieve this, we create a key frame video by incorporating the first frame of each short video segment to provide global guidance directly. ControlVideo is then applied to this key frame video, which is subsequently fused with the previously obtained $\hat{\vepsilon}_\theta$. Formally, let $X_t^K=\{\vx_t^{(j-1)(L-a)+1}\}_{j=1}^n$ denote the keyframe video and $C^K=\{c^{(j-1)(L-a)+1}\}_{j=1}^n$ denote the corresponding visual conditions. The final model $\vepsilon_\theta$ is defined as follows:
\begin{align}
\label{eq: key fusion}
    \vepsilon_\theta = wO(\epsilon_\theta^K) + (1-w)\hat{\vepsilon}_\theta,
\end{align}
where $w\in [0,1]$ is the weight, $\epsilon_\theta^K = \vepsilon(X_t^K, C^K, p, t)$. Note that the frames in keyframe videos here are also selected as key frames in each short video in key frame attention, thus ensuring global temporal consistency. The complete algorithm is presented in Algorithm \ref{alg: long video editing}. As depicted in Figure \ref{fig:ablations}, with the keyframe video fusion strategy, the color of the car is consistently retained throughout the entire video.

\begin{figure}
  \centering
  \includegraphics[width=1.0\columnwidth]{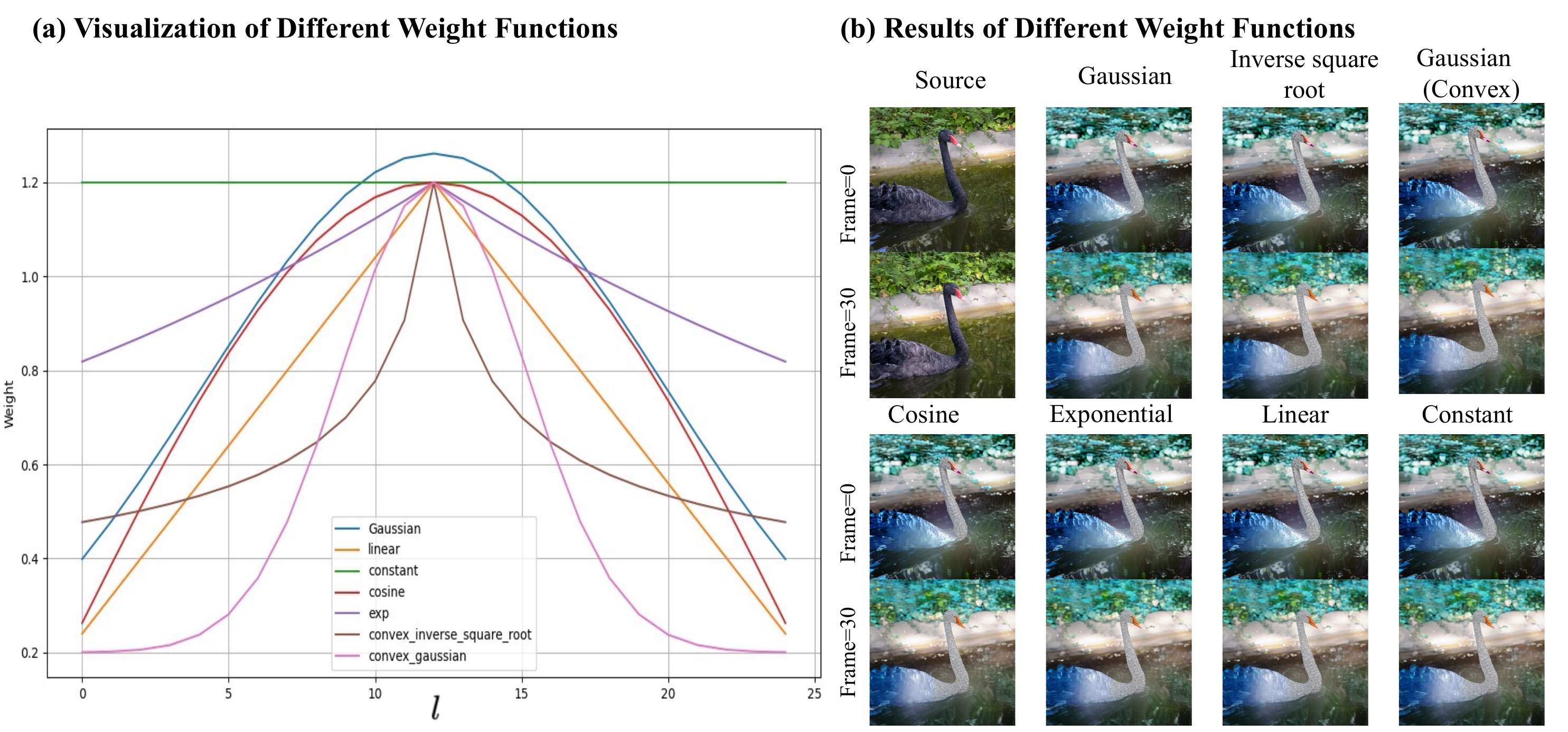}
  \caption{(a) Visualization of different weight functions, where we take $L=25$ as example. (b) The edited results with different weight functions.}
  \label{fig:weight}
\end{figure}

\section{Related Work}
\subsection{Diffusion Models for Text-driven Generation and Image Editing} 
Recently, diffusion models have achieved major breakthroughs in the field of generative artificial intelligence and thus are utilized for text-to-image generation ~\cite{rombach2022high, saharia2022photorealistic}. These models usually train a diffusion model conditioned on text on large-scale image-text paired datasets. Building upon these remarkable advances of T2I diffusion models, numerous methods have shown promising results in text-driven image editing. In particular, several works such as Prompt-to-Prompt~\cite{hertz2022prompt}, Plug-and-Play~\cite{tumanyan2022plug} and Pix2pix-Zero~\cite{parmar2023zero} explore the attention control over the generated content and achieve SOTA results. Such methods usually start from the DDIM inversion and replace attention maps in the generation process with the attention maps from the source prompt, which retrain the spatial layout of the source image. Despite significant advances, directly applying these image editing methods to video frames leads to temporal flickering.

\subsection{Diffusion Models for Text-driven Video Editing} Gen-1~\cite{esser2023structure} trains a video diffusion model on large-scale datasets, achieving impressive performance. However, it requires expensive computational resources. To overcome this, recent works build upon T2I diffusion models on a single text-video pair. In particular, Tune-A-Video~\cite{wu2022tune} inflates the T2I diffusion model to the T2V diffusion model and finetunes it on the source video-text data. Inspired by this, several works~\cite{qi2023fatezero,liu2023video,wang2023zero} combine it with attention map injection methods, achieving superior performance. Despite advances, empirical evidence suggests that they still struggle to faithfully and adequately control the output while preserving temporal consistency.
\begin{figure}
  \centering 
  \includegraphics[width=1.0\columnwidth]{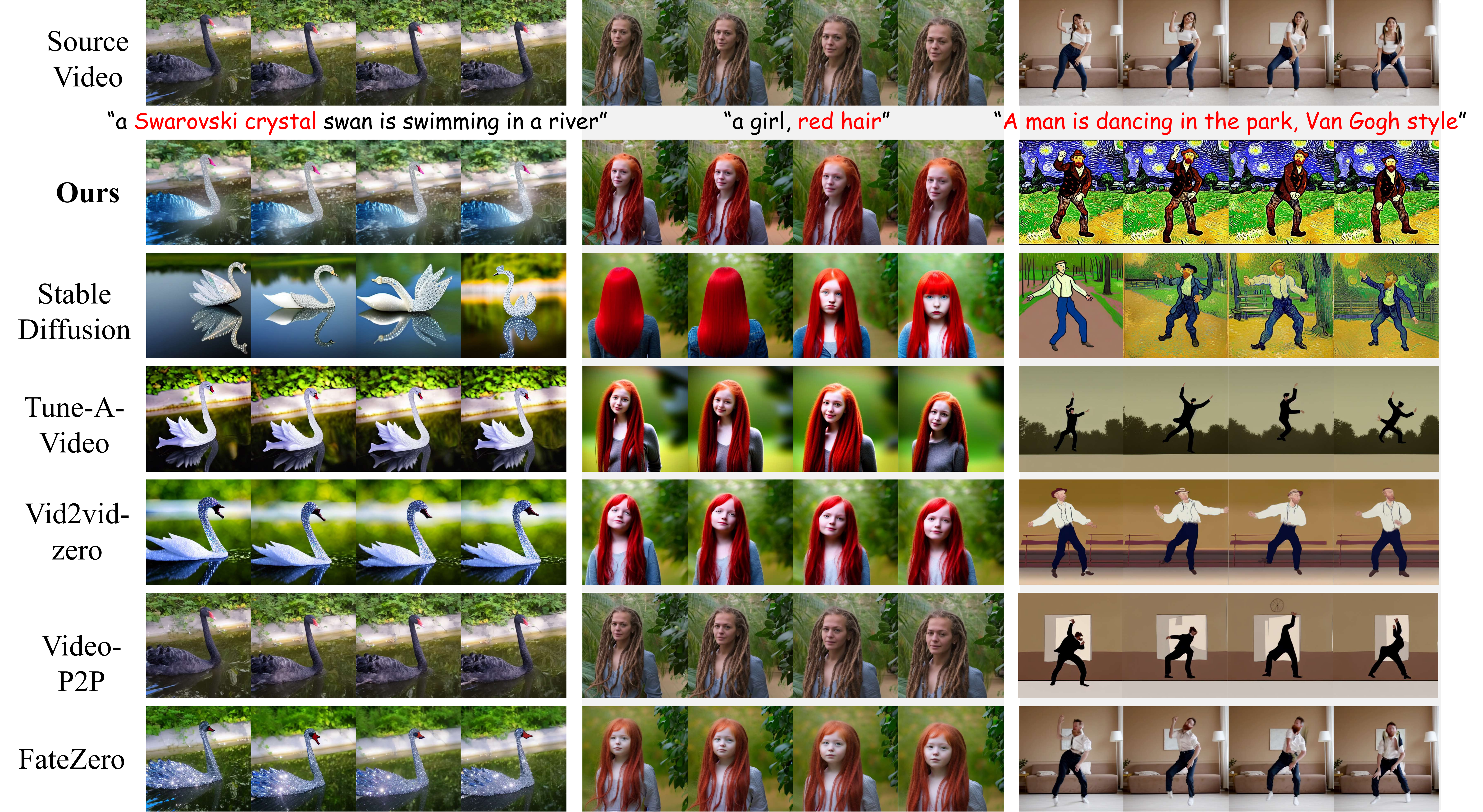}
  \caption{Comparison with baselines on DAVIS and collected data from the website. ControlVideo achieves better visual quality by fulfilling three requirements simultaneously. By starting from Gaussian noise rather than DDIM inversion, we can improve editability in global editing (see the third example). }
  \label{fig:baseline}
\end{figure}
\section{Experiments}
\subsection{Setup}
\label{sec: setup}
\subsubsection{Implementation Details} 
For short video editing, following previous research~\cite{wang2023zero}, we use 8 frames with $512\times512$ resolution for fair comparisons. We collect 50 video-text pair data from DAVIS dataset~\cite{pont20172017} and website\footnote{https://www.pexels.com}. We compare ControlVideo with Stable Diffusion and the following SOTA text-driven video editing methods: Tune-A-Video~\cite{wu2022tune}, Vid2vid-zero~\cite{parmar2023zero}, Video-P2P~\cite{liu2023video} and FateZero~\cite{qi2023fatezero}. By default, we train the ControlVideo for 80, 300, 500, and 1500 iterations for canny edge maps, HED boundary, depth maps, and pose respectively with a learning rate $3\times 10 ^{-5}$. The control scale $\lambda$ is set to 1. For multiple controls, we set $\lambda_i=0.5$ by default. The DDIM sampler~\cite{song2020denoising} with 50 steps and 12 classifier-free guidance are used for inference. The Stable Diffusion 1.5~\cite{rombach2022high} and ControlNet 1.0~\cite{zhang2023adding} with canny edge maps, HED boundary, depth maps, and pose are adopted in the experiment. For image-driven video editing, we employ the Lora weight from Civitai and merge it into Stable Diffusion. 

\subsubsection{Evaluation} 
Following the previous work~\cite{qi2023fatezero}, we report CLIP-temp for temporal consistency and CLIP-text for text alignment. We also report SSIM~\cite{wang2004image} within the unedited area between input-output pairs for faithfulness. The metric for faithfulness only considers the unedited area. The unedited area is computed by SAM~\cite{kirillov2023segany} according to text. Additionally, we perform a user study to quantify text alignment, temporal consistency, faithfulness, and overall all aspects by pairwise comparisons between the baselines and ControlVideo. A total of 10 subjects participated in this section. Taking faithfulness as an example, given a source video, the participants are instructed to select which edited video is more faithful to the source video in the pairwise comparisons between the baselines and ControlVideo.

\begin{figure}
  \centering
  \includegraphics[width=1.0\columnwidth]{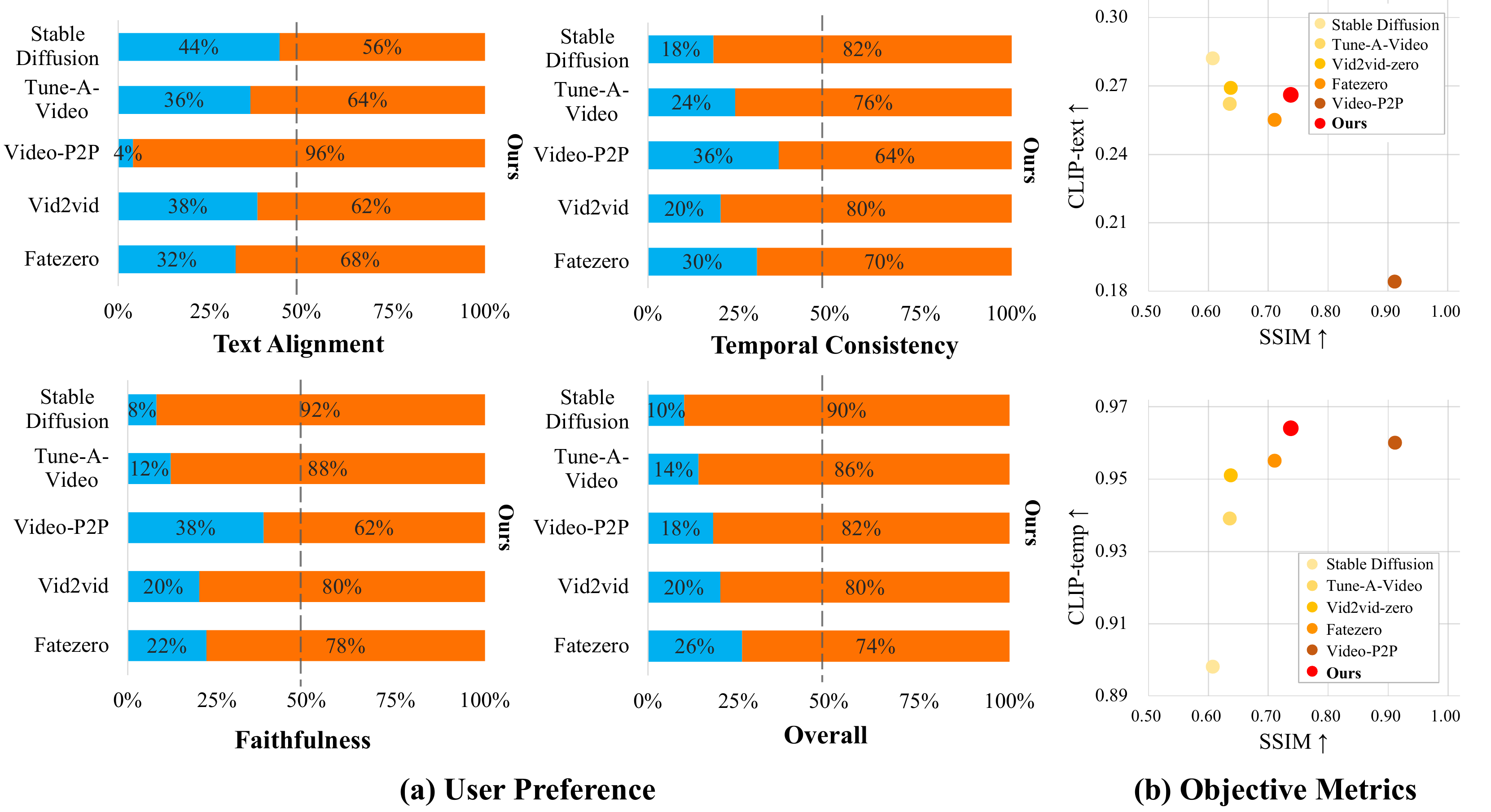}
  \caption{Quantitative results under user study and objective metrics. ControlVideo outperforms all baselines from overall aspects. See detailed analysis in Sec. \ref{sec: comparisons}.}
  \label{fig:quantative}
\end{figure}

\subsection{Results}
\subsubsection{Applications} 
\label{sec: apllications}
The main results are shown in Figure \ref{fig: main results}. Firstly, under the guidance of different \emph{single controls}, ControlVideo delivers videos with high visual realism in attributes, style, and background editing. For instance, HED boundary control helps to change the swan into a Swarovski crystal swan faithfully. Pose control allows shape modification flexibly by changing the man into Sherlock Holmes with a black coat. Secondly, in the ``person'' $\to$ ``panda'' case, ControlVideo can preserve the background and change the shape simultaneously by combining \emph{multiple controls} (Canny edge maps and pose control) to utilize the advantage of different control types. Moreover, in \emph{image-driven video editing}, ControlVideo successfully changes the woman in the source video into Evangeline Lilly to align the reference images. Finally, we can preserve the identity of the woman across hundreds of frames, demonstrating the ability of ControlVideo to maintain \emph{long-range temporal consistency}.

\subsubsection{Comparisons}
\label{sec: comparisons}
The quantitative and qualitative results are shown in Figure \ref{fig:quantative}  and Figure \ref{fig:baseline} respectively. We emphasize that text-driven video editing should fulfill three requirements simultaneously and a single objective metric cannot reflect the edited results. For instance, Video-P2P with high SSIM tends to reconstruct the source video and fails to align the text. As shown in Figure \ref{fig:baseline}, in the "a girl with red hair" example, it cannot change the hair color. Stable Diffusion and Vid2vid-zero with high CLIP-text generate a girl with striking red hair, but entirely ignore the identity of the female from the source video, leading to unsatisfactory results. 

As shown in Figure \ref{fig:quantative}(a), for overall aspects conducted by user study, our method outperforms all baselines significantly. Specifically, $86\%$ persons prefer our edited videos to Tune-A-Video. What's more, human evaluation is the most reasonable quantitative metric for video editing tasks and we can observe ControlVideo outperforms all baselines in all aspects. The qualitative results in Figure \ref{fig:baseline} are consistent with quantitative results, where ControlVideo not only successfully changes the hair color but also keeps the identity of the female unchanged while all existing methods fail. Overall, extensive results demonstrate that ControlVideo outperforms all baselines by delivering temporal consistent, and faithful videos while still aligning with the text prompt. 

\subsubsection{Ablation Studies for Key Components in ControlVideo}
\label{sec: ablations} 
As shown in Figure \ref{fig:ablations}, adding controls provides additional guidance from the source video, thus improving faithfulness a lot. The key-frame attention improves temporal consistency a lot. The temporal attention improves faithfulness and temporal consistency. Combining all the modules achieves the best performance. The quantitative results in the Appendix are consistent with the qualitative results.

\subsubsection{Ablation Studies for hyper-parameters in Long Video Editing}
\label{sec: ablations for long video} 
In this section, we perform ablation studies for overlapping Length $a$, weight $w$ for key frame video fusion and weight functions for fusion with nearby videos in extended controlvideo. As depicted in Figure \ref{fig:ablations}, an increased overlapping length $a$ yields videos with enhanced temporal consistency. In this study, we set $a \in [\frac{L}{2}, L]$. A larger $w$ promotes consistency over extended temporal sequences in whole videos. Nonetheless, too large $w$ can introduce temporal flickering.  In this work, we set $w \in [0.2, 0.5]$. Additionally, we devise a variety of weight functions for fusion with nearby videos. Given that fusion occurs at both ends of a video, we prefer to create functions that are symmetric about $L/2$ and maintain all elements greater than zero. As depicted in Figure \ref{fig:weight}, we explore several functional forms, including constant, linear, concave (e.g., cosine), and convex (e.g., inverse square root) functions. The outcomes presented in Figure \ref{fig:weight} indicate that the quality of the edited video remains largely unaffected by the choice of weight function.

\section{Conclusion}
In this paper, we present ControlVideo, a general framework to utilize T2I diffusion models for one-shot video editing, which incorporates additional conditions such as edge maps, the key frame and temporal attention to improve faithfulness and temporal consistency. We demonstrate its effectiveness by outperforming state-of-the-art text-driven video editing methods.

\bibliographystyle{unsrt}
\bibliography{example.bib}

\begin{thebibliography}{10}

\bibitem{qi2023fatezero}
Chenyang Qi, Xiaodong Cun, Yong Zhang, Chenyang Lei, Xintao Wang, Ying Shan, and Qifeng Chen.
\newblock Fatezero: Fusing attentions for zero-shot text-based video editing.
\newblock {\em arXiv preprint arXiv:2303.09535}, 2023.

\bibitem{wang2023zero}
Wen Wang, Kangyang Xie, Zide Liu, Hao Chen, Yue Cao, Xinlong Wang, and Chunhua Shen.
\newblock Zero-shot video editing using off-the-shelf image diffusion models.
\newblock {\em arXiv preprint arXiv:2303.17599}, 2023.

\bibitem{wu2022tune}
Jay~Zhangjie Wu, Yixiao Ge, Xintao Wang, Weixian Lei, Yuchao Gu, Wynne Hsu, Ying Shan, Xiaohu Qie, and Mike~Zheng Shou.
\newblock Tune-a-video: One-shot tuning of image diffusion models for text-to-video generation.
\newblock {\em arXiv preprint arXiv:2212.11565}, 2022.

\bibitem{liu2023video}
Shaoteng Liu, Yuechen Zhang, Wenbo Li, Zhe Lin, and Jiaya Jia.
\newblock Video-p2p: Video editing with cross-attention control.
\newblock {\em arXiv preprint arXiv:2303.04761}, 2023.

\bibitem{rombach2022high}
Robin Rombach, Andreas Blattmann, Dominik Lorenz, Patrick Esser, and Bj{\"o}rn Ommer.
\newblock High-resolution image synthesis with latent diffusion models.
\newblock In {\em Proceedings of the IEEE/CVF Conference on Computer Vision and Pattern Recognition}, pages 10684--10695, 2022.

\bibitem{ho2022imagen}
Jonathan Ho, William Chan, Chitwan Saharia, Jay Whang, Ruiqi Gao, Alexey Gritsenko, Diederik~P Kingma, Ben Poole, Mohammad Norouzi, David~J Fleet, et~al.
\newblock Imagen video: High definition video generation with diffusion models.
\newblock {\em arXiv preprint arXiv:2210.02303}, 2022.

\bibitem{hertz2022prompt}
Amir Hertz, Ron Mokady, Jay Tenenbaum, Kfir Aberman, Yael Pritch, and Daniel Cohen-Or.
\newblock Prompt-to-prompt image editing with cross attention control.
\newblock {\em International Conference on Learning Representations}, 2023.

\bibitem{tumanyan2022plug}
Narek Tumanyan, Michal Geyer, Shai Bagon, and Tali Dekel.
\newblock Plug-and-play diffusion features for text-driven image-to-image translation.
\newblock {\em arXiv preprint arXiv:2211.12572}, 2022.

\bibitem{parmar2023zero}
Gaurav Parmar, Krishna~Kumar Singh, Richard Zhang, Yijun Li, Jingwan Lu, and Jun-Yan Zhu.
\newblock Zero-shot image-to-image translation.
\newblock {\em arXiv preprint arXiv:2302.03027}, 2023.

\bibitem{zhang2023adding}
Lvmin Zhang and Maneesh Agrawala.
\newblock Adding conditional control to text-to-image diffusion models.
\newblock {\em arXiv preprint arXiv:2302.05543}, 2023.

\bibitem{hu2021lora}
Edward~J Hu, Yelong Shen, Phillip Wallis, Zeyuan Allen-Zhu, Yuanzhi Li, Shean Wang, Lu~Wang, and Weizhu Chen.
\newblock Lora: Low-rank adaptation of large language models.
\newblock {\em arXiv preprint arXiv:2106.09685}, 2021.

\bibitem{radford2021learning}
Alec Radford, Jong~Wook Kim, Chris Hallacy, Aditya Ramesh, Gabriel Goh, Sandhini Agarwal, Girish Sastry, Amanda Askell, Pamela Mishkin, Jack Clark, et~al.
\newblock Learning transferable visual models from natural language supervision.
\newblock In {\em International conference on machine learning}, pages 8748--8763. PMLR, 2021.

\bibitem{song2020score}
Yang Song, Jascha Sohl-Dickstein, Diederik~P Kingma, Abhishek Kumar, Stefano Ermon, and Ben Poole.
\newblock Score-based generative modeling through stochastic differential equations.
\newblock In {\em International Conference on Learning Representations}, 2020.

\bibitem{bao2021analytic}
Fan Bao, Chongxuan Li, Jun Zhu, and Bo~Zhang.
\newblock Analytic-dpm: an analytic estimate of the optimal reverse variance in diffusion probabilistic models.
\newblock In {\em International Conference on Learning Representations}, 2021.

\bibitem{ho2020denoising}
Jonathan Ho, Ajay Jain, and Pieter Abbeel.
\newblock Denoising diffusion probabilistic models.
\newblock {\em Advances in Neural Information Processing Systems}, 33:6840--6851, 2020.

\bibitem{song2020denoising}
Jiaming Song, Chenlin Meng, and Stefano Ermon.
\newblock Denoising diffusion implicit models.
\newblock {\em arXiv preprint arXiv:2010.02502}, 2020.

\bibitem{meng2021sdedit}
Chenlin Meng, Yang Song, Jiaming Song, Jiajun Wu, Jun-Yan Zhu, and Stefano Ermon.
\newblock Sdedit: Image synthesis and editing with stochastic differential equations.
\newblock {\em International Conference on Learning Representations}, 2022.

\bibitem{zhao2022egsde}
Min Zhao, Fan Bao, Chongxuan Li, and Jun Zhu.
\newblock Egsde: Unpaired image-to-image translation via energy-guided stochastic differential equations.
\newblock {\em Advances in Neural Information Processing Systems}, 35:3609--3623, 2022.

\bibitem{kirillov2023segany}
Alexander Kirillov, Eric Mintun, Nikhila Ravi, Hanzi Mao, Chloe Rolland, Laura Gustafson, Tete Xiao, Spencer Whitehead, Alexander~C. Berg, Wan-Yen Lo, Piotr Doll{\'a}r, and Ross Girshick.
\newblock Segment anything.
\newblock {\em arXiv:2304.02643}, 2023.

\bibitem{singer2022make}
Uriel Singer, Adam Polyak, Thomas Hayes, Xi~Yin, Jie An, Songyang Zhang, Qiyuan Hu, Harry Yang, Oron Ashual, Oran Gafni, et~al.
\newblock Make-a-video: Text-to-video generation without text-video data.
\newblock {\em arXiv preprint arXiv:2209.14792}, 2022.

\bibitem{jimenez2023mixture}
{\'A}lvaro~Barbero Jim{\'e}nez.
\newblock Mixture of diffusers for scene composition and high resolution image generation.
\newblock {\em arXiv preprint arXiv:2302.02412}, 2023.

\bibitem{saharia2022photorealistic}
Chitwan Saharia, William Chan, Saurabh Saxena, Lala Li, Jay Whang, Emily~L Denton, Kamyar Ghasemipour, Raphael Gontijo~Lopes, Burcu Karagol~Ayan, Tim Salimans, et~al.
\newblock Photorealistic text-to-image diffusion models with deep language understanding.
\newblock {\em Advances in Neural Information Processing Systems}, 35:36479--36494, 2022.

\bibitem{esser2023structure}
Patrick Esser, Johnathan Chiu, Parmida Atighehchian, Jonathan Granskog, and Anastasis Germanidis.
\newblock Structure and content-guided video synthesis with diffusion models.
\newblock {\em arXiv preprint arXiv:2302.03011}, 2023.

\bibitem{pont20172017}
Jordi Pont-Tuset, Federico Perazzi, Sergi Caelles, Pablo Arbel{\'a}ez, Alex Sorkine-Hornung, and Luc Van~Gool.
\newblock The 2017 davis challenge on video object segmentation.
\newblock {\em arXiv preprint arXiv:1704.00675}, 2017.

\bibitem{wang2004image}
Zhou Wang, Alan~C Bovik, Hamid~R Sheikh, and Eero~P Simoncelli.
\newblock Image quality assessment: from error visibility to structural similarity.
\newblock {\em IEEE transactions on image processing}, 13(4):600--612, 2004.

\end{thebibliography}

\end{document}